\pdfoutput=1
\documentclass[sigconf,nonacm]{acmart}

\usepackage{booktabs}
\usepackage{amsmath}
\usepackage{graphicx}

\usepackage{microtype}
\usepackage[htt]{hyphenat}
\begin{document}

\title{The Split Protocol Decides the Result:
       Re-Evaluating Encrypted Traffic Classification}

\author{Mayan Sharma}
\affiliation{%
  \institution{Vellore Institute of Technology}
  \city{Chennai}
  \country{India}}

\author{Kriti Saini}
\affiliation{%
  \institution{Vellore Institute of Technology}
  \city{Chennai}
  \country{India}}

\author{Devansh Behl}
\affiliation{%
  \institution{Vellore Institute of Technology}
  \city{Chennai}
  \country{India}}

\renewcommand{\shortauthors}{Sharma, Saini and Behl}



\begin{abstract}
Encrypted traffic classifiers are routinely reported at 95--99\% accuracy. We
show that on one of two public corpora, the same model and the same code score
macro-F1 \textbf{0.992} under the evaluation protocol standard in the
literature and \textbf{0.567} under a protocol that keeps flows from one packet
capture on one side of the split. Nothing changes but the partition. The gap is
$0.426$ macro-F1 over eight seeds, $p = 0.008$ after Holm-Bonferroni
correction.

The inflation is not a property of random splitting as such. On a second
corpus, a month of backbone QUIC traffic, we evaluate five split protocols at
eight seeds each: random, temporal and \emph{client}-disjoint splitting are all
statistically indistinguishable from session-disjoint, with $|d| < 0.35$ and
$p > 0.5$. What costs $0.192$ macro-F1 there is \emph{server}-disjoint
splitting ($d = 17.0$, $p = 0.008$). \textbf{Four of five axes are null and one
is decisive}, and which one depends on how the corpus was collected in a way
invisible from its label space. Controlling the wrong axis protects nothing.

We give a diagnostic that identifies the right axis before any model is
trained: single-identifier probes. On the first corpus a decision tree on the
\emph{server address alone} reaches 0.999 under a random split and the
\emph{capture identifier alone} reaches 0.983; under session-disjoint splitting
the latter falls to exactly 0.000. On the second, the \emph{SNI string alone}
reaches 0.967 under every split, which is the ceiling of the labelling process
rather than of any classifier.

Evaluated this way, we compare a metric-trained flow embedding against eight
classical and four deep baselines on identical committed splits. It leads every
deep baseline and does not beat gradient-boosted trees on twenty packet sizes.
Five of our own architectural claims fail their own pre-registered tests,
including the one we expected to survive: distance-based rejection of unknown
applications beats softmax thresholding by a wide margin, but ablating
\emph{every} metric objective makes rejection \emph{better}. Measuring the
embedding geometry explains why --- class separation correlates at $r = -0.88$
with rejection quality, so an objective optimising for discrimination among
known classes works against detecting unknown ones.

A sixth claim was mis-specified rather than false. Few-shot enrollment is flat
within a corpus, where the prototypes are already correct; \emph{across}
corpora, five labelled flows per class recover 91\% of the deficit left by
failed zero-shot transfer.

All splits, results, code and negative findings are released.
\end{abstract}


\maketitle

\section{Introduction}
\label{sec:introduction}

Encrypted traffic classification has a reproducibility problem that is easy to
state. Papers report macro-F1 and accuracy above 95\%, sometimes above 99\%,
on datasets assembled from packet captures. Those numbers are real in the sense
that the code produces them. They are not results about traffic analysis.

We demonstrate this directly. Table~\ref{tab:split-contrast} reports the same
model, the same code, the same hyperparameters and the same data, evaluated
under four ways of partitioning the flows into training and test sets. On the
5G Traffic corpus the model scores macro-F1 \textbf{0.992} under a stratified
random split over flows --- the protocol used, to our reading, by the large
majority of published work in this area --- and \textbf{0.567} when flows from
one packet capture are constrained to a single side of the split. A 2016
statistical baseline behaves identically: 0.996 and 0.640.

The obvious diagnosis, that random splitting inflates results, is wrong, or at
least incomplete. On CESNET-QUIC22, a month of QUIC traffic from a national
backbone, we evaluated five split protocols at eight seeds each. Random,
temporal and client-disjoint splitting are statistically indistinguishable from
session-disjoint --- $p > 0.5$ and $|d| < 0.35$ in all three cases, which are
powered nulls rather than absences of evidence. What costs
\textbf{0.192 macro-F1} there is \emph{server}-disjoint splitting: testing only
on flows to servers never seen in training ($d = 17.0$, $p = 0.0078$, surviving
Holm-Bonferroni correction across the family).

\subsection*{The axis that matters is corpus-specific}

The two corpora fail in different places because they were collected
differently. 5G Traffic is a set of single-application packet captures: one
recording, one app, one device. A flow's capture is therefore nearly its label,
and a classifier that identifies the capture scores as well as one that
identifies the application. CESNET-QUIC22 is backbone traffic from many hosts
to many servers; its flows are already independent across days, so grouping by
day buys nothing, but the label correlates strongly with the destination, so
holding out servers bites.

Neither protocol protects against the other's failure. A paper reporting
session-disjoint results on backbone traffic has controlled nothing. A paper
reporting server-disjoint results on single-session captures has controlled
nothing either.

\subsection*{A diagnostic that runs before the model}

The right axis can be identified without training anything. We fit a decision
tree on a \emph{single} identifier column and report its macro-F1 under each
candidate protocol. On 5G Traffic the server address alone reaches 0.999 under
a random split and the capture-session identifier alone reaches 0.983; under
session-disjoint splitting the latter falls to exactly 0.000, which is the
grouping doing precisely what it claims. On CESNET-QUIC22 the SNI string alone
reaches 0.967 under every split --- unsurprising, since that dataset's labels
are derived from SNI, and it therefore measures the ceiling of the labelling
process rather than of any classifier.

These probes cost seconds. We argue they should precede any reported number in
this area, and we release them as a package that runs on any corpus in our
schema.

\subsection*{What survives under strict evaluation}

Having established the protocol, we use it. We evaluate a metric-trained
flow-level embedding against eight classical and four deep baselines on
identical committed splits, and we report five of our own architectural claims
failing their own pre-registered tests. Most notably, distance-based rejection
of unknown applications beats softmax thresholding by a wide margin --- 0.34
against 0.61 false-accept rate at matched true-positive rate --- but ablating
\emph{every} metric objective from the training makes rejection \emph{better},
not worse. The advantage belongs to the rejection rule, not to the
representation learning we introduced to produce it. We report this because a
reviewer would otherwise find it, and because the corrected claim is more
useful: it applies to anyone's embedding.

\subsection*{Contributions}

\begin{figure}[t]
\centering
\includegraphics[width=\columnwidth]{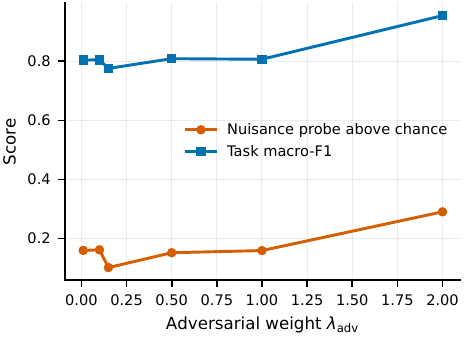}
\caption{Nuisance leakage and task performance as the adversarial weight
varies. The trade-off curve this figure was designed to show does not exist:
leakage is flat across two orders of magnitude in $\lambda$ and sits at or above
the raw-feature control throughout. The figure is included because a flat line
here is the result.}
\label{fig:adversarial}
\end{figure}

\begin{enumerate}
\item A measurement of how much the split protocol determines the reported
      result, on two corpora with opposite structure, every method on identical
      committed splits and every protocol at eight seeds
      (\S\ref{sec:protocols}). The same change is worth $0.426$ macro-F1 on one
      corpus and $0.001$ on the other. On the second, four of five axes are
      statistically null and the fifth costs $0.192$; the positive results
      survive Holm-Bonferroni correction and the nulls are powered rather than
      undetermined.
\item Single-identifier probes as a pre-training diagnostic for which axis a
      corpus requires, with the mechanism made explicit
      (\S\ref{sec:protocols}).
\item An evaluation of thirteen methods under strict protocols
      (\S\ref{sec:results}), including the finding that on a volume-defined
      taxonomy four independent neural architectures fall within $0.05$ macro-F1
      of each other while gradient-boosted trees on thirteen flow statistics sit
      $0.23$ above all of them, and an early-decision curve
      (Figure~\ref{fig:early}) and countermeasure-overhead evaluation
      (Figure~\ref{fig:robustness}) under the same protocols.
\item Five negative architectural results, each with the test that produced it
      (\S\ref{sec:negative}), including a representation-probing protocol
      showing an adversarial component leaving its nuisance as decodable as the
      raw input at every weight (Figure~\ref{fig:adversarial}), and a geometric
      explanation for why ablating our metric objectives \emph{improves}
      open-set rejection.
\item A sixth claim corrected rather than withdrawn: enrollment is flat within
      a corpus and recovers 91\% of a failed zero-shot transfer across corpora
      (\S\ref{sec:results}, Figure~\ref{fig:enrollment}).
\item An artifact containing every split manifest, every result including the
      failed experiments, the audit that found them, and a two-minute pipeline
      verification requiring no data download (\S\ref{sec:threats}).
\end{enumerate}

\subsection*{What we do not claim}

We do not claim state-of-the-art accuracy. On the corpus where our model is
strongest it is within one seed's noise of gradient-boosted trees on twenty
packet sizes, which train in three seconds. We did not run ET-BERT or the other
pre-trained byte-level transformers, because they tokenise payload bytes that
QUIC does not expose and our records do not retain; we say so rather than
citing their published numbers alongside ours.

\section{Background and Related Work}
\label{sec:background}

\subsection{Classical fingerprinting}

Traffic analysis without payload inspection has a long line of work built on
packet sizes, directions and inter-arrival times. CUMUL~\cite{panchenko2016website}
interpolates the cumulative signed-size curve to a fixed-length vector.
AppScanner~\cite{taylor2016appscanner} computes eighteen summary statistics over
the incoming, outgoing and combined size series and fits a random forest.
$k$-fingerprinting~\cite{hayes2016kfingerprinting} adds burst structure and
concentration features. All three predate the deep-learning literature in this
area by several years, and we find in \S\ref{sec:results} that AppScanner
remains within 0.02 macro-F1 of a modern neural model on one of our corpora.

\subsection{Neural classifiers}

Deep Packet~\cite{lotfollahi2020deeppacket} applies a 1D convolutional network
to the first 1500 payload bytes. FS-Net~\cite{liu2019fsnet} encodes discretised
packet lengths with a stacked bidirectional GRU. More recently, pre-trained
transformers such as ET-BERT~\cite{lin2022etbert} tokenise raw datagram bytes
and report macro-F1 above 98\% on several standard benchmarks.

\paragraph{Why we do not compare against ET-BERT.} Its input is the hexadecimal
byte string of the packet payload, tokenised as bigrams over a 65{,}536-entry
vocabulary. Our corpora are QUIC and TLS, where the payload is encrypted, and
our records retain sizes, directions and inter-arrival times but no bytes.
Feeding it header-derived features would not be ET-BERT; it would be a
differently-sized transformer carrying the name. Its published pre-training
corpus is also not redistributed. We therefore do not report a number for it,
rather than reporting one we could not obtain fairly. The same reasoning
applies to YaTC, CLE-TFE, MIETT, FlowletFormer and PacketCLIP; the artifact
documents each case individually.

This is a real limitation and it bounds our claims: we do not assert
state-of-the-art accuracy, and \S\ref{sec:results} does not contain the
strongest published models.

\subsection{Evaluation protocol}

The methodological question we raise is not new in machine learning generally.
Group-aware cross-validation is standard wherever observations cluster, and
tree-based models are known to outperform neural networks on tabular
features~\cite{grinsztajn2022trees} --- a result we reproduce in
\S\ref{sec:results} in a setting where it is rarely tested.

What is specific to encrypted traffic is \emph{which} grouping matters, and
that this differs by corpus in a way that is not visible from the label space.
A dataset of per-application packet captures and a dataset of backbone flows
can carry the same label taxonomy, the same feature schema and the same class
balance, and require entirely different partitions to be evaluated honestly.
We are not aware of prior work that measures this contrast directly on two
corpora, or that provides a pre-training diagnostic for choosing the axis.

\subsection{Open-set recognition}

Deployments encounter applications absent from training. Rejection by maximum
softmax probability is the usual baseline; energy scores~\cite{liu2020energy}
and Mahalanobis distance to class-conditional Gaussians are common
alternatives. We evaluate all of these on a single frozen embedding
(\S\ref{sec:results}), which isolates the scoring rule from the representation
and produces the negative result in \S\ref{sec:negative}: the rejection
advantage we measured is a property of the rule, not of the metric training we
added to produce it.

\section{Datasets}
\label{sec:datasets}

We use two public corpora with deliberately opposite collection structure. They
are processed independently and never merged; \S\ref{sec:threats} explains why
merging them was itself a source of leakage in an earlier version of this work.

\subsection{5G Traffic}

Seventy-five Wireshark packet captures across fifteen applications in six
service categories, recorded on a commercial 5G network. One capture contains
one application. We read all 330 million packet rows directly from the
distributed archive and extract \emph{conversation segments}: packets between
the capture's client host and one server over one transport, bounded by a
10-second idle timeout and a 10-second active timeout. This yields
164{,}348 flows.

The service label is the dataset's own directory taxonomy
(\texttt{Game\_Streaming}, \texttt{Online\_Game}, \texttt{Live\_Streaming},
\texttt{Stored\_Streaming}, \texttt{Metaverse}, \texttt{Video\_Conferencing}),
so no hand-written mapping intervenes between the collectors' intent and our
labels.

\paragraph{The structural property that matters.} Several applications have only
one or two capture files. Grouping by capture therefore places an entire
application on one side of the split: in our seed-42 session-disjoint partition
the whole of \texttt{roblox} lands in test while training sees only
\texttt{zepeto} for the \texttt{metaverse} class. The 5G session-disjoint
number is consequently a hybrid of session- and application-disjointness. We
report it as such rather than presenting it as a clean session-disjoint result.

\subsection{CESNET-QUIC22}

Twenty-eight daily flow files spanning four consecutive weeks of QUIC traffic
from a national research backbone~\cite{luxemburk2023cesnet}. Each record is
already a bidirectional flow with the first thirty packets as
(inter-arrival, direction, size) triples, plus addresses, ports, destination AS
and the SNI string. We scan all 153 million rows and retain a seeded reservoir
of 400 flows per service category per day, giving 201{,}600 flows across
eighteen categories, perfectly balanced by construction.

\paragraph{Labels come from SNI.} The publishers derive the application and
category fields from the TLS/QUIC server name. Our task is therefore
\emph{predict what SNI would have said, without seeing SNI} --- the right task
for a world with Encrypted Client Hello~\cite{rescorla2024ech}, and one whose
ceiling is set by how well SNI itself categorises traffic. We quantify that
ceiling directly in \S\ref{sec:protocols}.

\subsection{A third corpus, prepared but not run}

Our claim that the split axis is corpus-specific rests on two datasets, which
is enough to show the axis differs between them and not enough to establish a
general rule. The artifact therefore includes a loader for ISCX-Tor2016,
selected over ISCX-VPN2016, USTC-TFC2016 and CIC-IoT2023 for one reason: it is
the only public corpus we found that offers a grouping axis which is
\emph{neither temporal nor per-file}. Its collection captured the same activity
simultaneously at the workstation and at the gateway and released the captures
in pairs, so \texttt{vantage} becomes a split axis and the protocol comparison
becomes three-way.

We had not obtained it at the time of writing --- it is distributed behind a
registration form --- so no number from it appears here. The loader is released
with every assumption it makes about the on-disk layout marked explicitly and
paired with the check that would falsify it, and the preparer warns if more
than 5\% of rows fail vantage inference, which is what a wrong assumption would
look like.

The other three were rejected for reasons worth recording. CIC-IoT2023's usable
CSV tier carries no addresses, ports or timestamps, so no grouped split is
possible on it, and its groupable tier is 548\,GB; it is also not encrypted
traffic. USTC-TFC2016 collected its benign classes with a traffic-simulation
appliance and its malware classes from real networks, so any split-protocol
result on it would be confounded with the generator. ISCX-VPN2016 offers one
grouping key rather than two, and ISCX-Tor2016's non-Tor half is recycled from
it, so the two are not independent corpora and must not both be used as though
they were.

\subsection{Schema and what the model may see}

Both corpora are reduced to one schema separating \emph{behaviour} from
\emph{provenance}. Behaviour --- per-packet sizes, directions and inter-arrival
times, plus flow-level aggregates --- is what a model may consume. Provenance
--- flow identifier, capture identifier, timestamp, client and server address,
destination port, SNI, AS number, transport protocol --- is retained but
excluded from model inputs, and exists so that the split protocols can group on
it and the probes of \S\ref{sec:protocols} can measure it.

The transport protocol number is provenance rather than input, which is not
obvious and is worth stating. In an earlier merged version of these corpora,
\texttt{protocol} took the value 0 on 11{,}985 rows of which 99.9\% belonged to
one class --- because 0 meant ``the preprocessing script could not parse this
field'', and only one script produced it. The column recorded which pipeline
had run. We now treat it as provenance and probe it like any other identifier.

A test suite asserts mechanically that no provenance column reaches a feature
extractor, and that every split is a partition with no identifier spanning
train and test. It runs in continuous integration and fails the build.

\section{Split Protocols and Identifier Probes}
\label{sec:protocols}

\subsection{Protocols}

We implement six partitions and report four. \textbf{Random flow} is stratified
random assignment of individual flows, the protocol we take to be standard in
this literature; we report it only as a contrast. \textbf{Session-disjoint}
assigns whole capture sessions --- a capture file for 5G Traffic, a capture day
for CESNET --- to exactly one side. \textbf{Temporal} sorts by first-packet
timestamp and takes the earliest flows for training. \textbf{Server-disjoint}
assigns whole server addresses to one side. Every protocol is materialised as a
committed manifest carrying per-side row indices and a SHA256 over the flow
identifiers, so a reader can verify that a reported number was produced on the
released partition.

\subsection{Identifier probes}

Before training anything, we fit a depth-limited decision tree on a
\emph{single} identifier column and score it under each protocol. The
identifiers are destination port, SNI string, server address and \texttt{/24}
prefix, destination AS, capture-session identifier, and transport protocol. A
probe whose column is absent or constant in a corpus reports
\emph{unavailable}, never a degenerate number --- a dataset with no SNI must
not be reported as one where SNI does not help.

\begin{table}[t]
\centering
\small
\begin{tabular}{lcc}
\toprule
Split protocol & cesnet\_quic22 & fiveg\_traffic \\
\midrule
\multicolumn{3}{l}{\emph{FlowCon-X (prototype head)}} \\
Random flow & 0.781 $\pm$ 0.006 & 0.992 $\pm$ 0.001 \\
Session-disjoint & 0.780 $\pm$ 0.006 & 0.567 $\pm$ 0.038 \\
Temporal & 0.781 $\pm$ 0.005 & 0.433 $\pm$ 0.080 \\
Server-disjoint & 0.588 $\pm$ 0.007 & 0.637 $\pm$ 0.010 \\
\midrule
\multicolumn{3}{l}{\emph{AppScanner (2016), identical splits}} \\
Random flow & 0.768 & 0.996 \\
Session-disjoint & 0.772 & 0.640 \\
Temporal & 0.761 & 0.255 \\
Server-disjoint & 0.566 & 0.711 \\
\bottomrule
\end{tabular}
\caption{Macro-F1 under four split protocols, every method on the identical committed splits. The random-flow row is the protocol standard in the literature and is reported only as a contrast: correlated flows from one capture session appear on both sides of it. \textbf{The size of that contrast is dataset-dependent}, which is itself the result: on a corpus of single-session captures a flow's capture is nearly its label, while on a backbone corpus whose flows are already independent the protocol makes little difference. The caption deliberately does not assert a direction; read the columns.}
\label{tab:split-contrast}
\end{table}

\subsection{What the probes find}

\paragraph{5G Traffic: the capture is the label.} Under a random split the
server address alone reaches macro-F1 0.999 and the capture-session identifier
alone 0.983. A classifier that learned nothing except which recording a flow
came from would score in the range published work reports for this corpus.
Under session-disjoint splitting the capture probe falls to \textbf{exactly
0.000} --- the grouping doing precisely what it claims --- and AppScanner falls
from 0.996 to 0.640, our model from 0.992 to 0.567. Temporal splitting is
harder still: 0.255 and 0.391 respectively. These captures span five months, so
a temporal split separates application versions and network conditions as well
as time, and it is the protocol under which this corpus behaves least like the
one the literature reports.

\paragraph{CESNET-QUIC22: the server is the label, and only the server.} We
evaluated five protocols at eight seeds each and tested each against
session-disjoint with a Wilcoxon signed-rank test, Holm-Bonferroni corrected
across the family:

\begin{center}
\small
\begin{tabular}{lrrr}
\toprule
Split protocol & Macro-F1 & Cohen's $d$ & $p$ \\
\midrule
Session-disjoint (ref.) & $0.7798 \pm 0.0061$ & --- & --- \\
Random flow & $0.7805 \pm 0.0057$ & $-0.12$ & 0.742 \\
Temporal & $0.7811 \pm 0.0051$ & $-0.31$ & 0.547 \\
Client-disjoint & $0.7791 \pm 0.0055$ & $0.13$ & 0.945 \\
\textbf{Server-disjoint} & $\mathbf{0.5881 \pm 0.0068}$ & $\mathbf{17.03}$ & $\mathbf{0.0078}^{*}$ \\
\bottomrule
\end{tabular}
\end{center}

\textbf{Four of five axes make no difference; the fifth costs 0.192 macro-F1.}
The three nulls are properly powered rather than absences of evidence: eight
seeds, effect sizes below 0.35, $p > 0.5$. Only server-disjointness survives
the correction ($^{*}$).

The client-disjoint row is worth its own sentence, because we added it
expecting the opposite. A third of the 48{,}072 distinct client addresses in
this corpus appear on more than one capture day, so grouping by day
demonstrably does not separate clients --- and separating them changes nothing
($p = 0.945$). Backbone flows from one client on different days are no more
alike, for this task, than flows from different clients.

AppScanner behaves the same way: 0.768, 0.772 and 0.762 under random,
session-disjoint and temporal splitting. Server-disjoint splitting takes it to 0.566 and our model from 0.781 to 0.588,
and takes the server-address probe from 0.772 to \textbf{0.026} --- the floor,
confirming the partition does what it says. Table~\ref{tab:split-contrast}
shows the two corpora side by side: the CESNET column is flat across three
protocols and drops only on the fourth, while the 5G column drops on all
three.

\paragraph{The labelling ceiling.} The SNI probe reaches 0.967 on CESNET under
every protocol except server-disjoint, where it still reaches 0.791. This is
expected, since that corpus's labels are derived from SNI, and it is the most
important number in this section: no result on CESNET-QUIC22 can be read as
matching ground truth, because the ground truth \emph{is} a name-based
heuristic and a decision tree recovers it at 0.967.

\subsection{The rule we propose}

The two corpora fail on different axes, and neither protocol protects against
the other's failure mode. We propose that the axis be chosen by measurement
rather than convention:

\begin{quote}
Before reporting any classifier result, fit a single-identifier probe for each
identifier the corpus retains, under each candidate protocol. Adopt the
protocol under which every identifier probe collapses toward the majority-class
floor. Report the probe table alongside the results.
\end{quote}

This costs seconds and it is falsifiable: if no protocol collapses the probes,
the corpus cannot support the claim being made of it. On 5G Traffic the rule
selects session-disjoint or temporal; on CESNET-QUIC22 it selects
server-disjoint. Adopting session-disjoint splitting on CESNET --- the choice a
practitioner following current best practice would make --- controls nothing
there, and our measurements say so.

\section{Model and Evaluation Setup}
\label{sec:model}

The classifier is a vehicle for the protocol comparison, and we describe it
briefly. Its architectural claims are examined --- and largely refuted --- in
\S\ref{sec:negative}.

\subsection{Architecture}

Two encoders read two views of a flow. The \emph{sequence encoder} consumes
eight per-packet channels: normalised size, log inter-arrival, direction,
signed size, log cumulative bytes, log cumulative time, size delta, and
position within the current same-direction run. All eight vary within a flow;
none is a flow-level constant broadcast across positions. It applies three
gated depthwise-separable convolutions and one transformer layer with
sinusoidal positions, then attention pooling. The \emph{context encoder} reads
ten flow-level aggregates through a small MLP. Cross-attention fuses them into
a 256-dimensional embedding $z_{\mathrm{flow}}$.

Training combines a supervised contrastive objective~\cite{khosla2020supcon}
with a pairwise margin term on $z_{\mathrm{flow}}$, a prototype alignment loss,
a cross-covariance disentanglement penalty between the two views, and a
gradient-reversal head~\cite{ganin2015grl} predicting a nuisance variable drawn
from provenance the model never sees. Every term is independently switchable
from configuration, which is what makes \S\ref{sec:negative} possible.

\paragraph{Causality of the cumulative channels.} The two cumulative channels
are normalised by fixed constants rather than by the observation window's
total. Normalising by the total would make the value at packet 1 depend on
packets not yet observed --- harmless for full-flow scoring, but it silently
inflates early-classification curves. A test asserts that truncating a sequence
is exactly equal to observing fewer packets.

\subsection{Evaluation}

Embeddings are scored with four heads on the same representation: cosine
$k$-NN, nearest class prototype, a logistic probe, and a linear SVM. Headline
metrics are macro-F1 and balanced accuracy with per-class F1 reported
alongside; a majority-class predictor scores 0.006 macro-F1 on CESNET, so
accuracy alone is uninformative. Macro averages exclude classes with zero test
support rather than scoring them as zero, and report how many were excluded ---
under 5G session-disjoint splitting, six of fifteen applications have no test
rows at all, and averaging them in as zeros would report a number about the
split rather than about the model.

\subsection{Reproducibility}

Every run seeds Python, NumPy and PyTorch, requests deterministic algorithms,
and records into its result file the seed, a hash of the resolved
configuration, the git commit and whether the working tree was dirty, library
versions, the device, and wall-clock time. We verified that deterministic
kernels are genuinely in force on our accelerator rather than silently falling
back: two same-seed runs produce bitwise-identical losses.

\paragraph{Budget, stated because it bounds the numbers.} Runs use fifteen
epochs at batch 512. Thirteen of sixteen headline runs ended with validation
macro-F1 still rising, so \textbf{every figure we report for our own model is a
lower bound}, and the comparison against classical baselines is unfair in our
own disfavour --- those are trained to their own convergence criteria. This is
a compute decision, not a modelling one, and it does not change any conclusion:
the largest gap we report is 0.30 macro-F1 and the per-epoch gain at epoch 15
is 0.008.

\section{Results Under Strict Protocols}
\label{sec:results}

Every method in this section is trained and evaluated on the identical
committed splits, with the same optimiser, epoch budget, class weighting and
early-stopping rule where those apply. Deviations from each baseline's original
configuration are recorded in the artifact and in the result files themselves,
never left unstated.

\begin{table}[t]
\centering
\small
\setlength{\tabcolsep}{4pt}
\begin{tabular}{lccc}
\toprule
Method & Macro-F1 & Balanced acc. & Params \\
\midrule
Majority class & 0.006 & 0.056 & --- \\
\midrule
\multicolumn{4}{l}{\emph{Identifier shortcuts}} \\
Destination port only & --- & --- & --- \\
SNI string only & 0.967 & 0.968 & --- \\
Server IP / prefix only & 0.768 & 0.769 & --- \\
Server AS only & 0.162 & 0.229 & --- \\
Transport protocol only & --- & --- & --- \\
\midrule
\multicolumn{4}{l}{\emph{Classical statistical}} \\
5 flow statistics (RF) & 0.415 & 0.421 & --- \\
All flow scalars (XGB) & 0.525 & 0.536 & --- \\
Packet-size histogram (XGB) & 0.713 & 0.719 & --- \\
First 10 packet sizes (XGB) & 0.721 & 0.726 & --- \\
First 20 packet sizes (XGB) & 0.790 & 0.793 & --- \\
CUMUL & 0.373 & 0.402 & --- \\
AppScanner & 0.772 & 0.775 & --- \\
$k$-fingerprinting & 0.590 & 0.597 & --- \\
\midrule
\multicolumn{4}{l}{\emph{Deep, non-pretrained}} \\
DeepPacket-style CNN & 0.651 $\pm$ 0.002 & 0.662 $\pm$ 0.002 & 270\,918 \\
FS-Net & 0.760 $\pm$ 0.005 & 0.765 $\pm$ 0.004 & 634\,258 \\
Bi-LSTM + attention & 0.705 $\pm$ 0.003 & 0.713 $\pm$ 0.003 & 573\,331 \\
Flow-statistics MLP & 0.260 $\pm$ 0.004 & 0.298 $\pm$ 0.005 & 73\,234 \\
\midrule
\textbf{FlowCon-X} & \textbf{0.780 $\pm$ 0.006} & 0.785 $\pm$ 0.006 & 1\,402\,165 \\
\bottomrule
\end{tabular}
\caption{Main comparison on cesnet\_quic22 under the session-disjoint protocol, mean $\pm$ std over seeds, every method on the identical committed splits. Identifier shortcuts are included deliberately: if one of them approaches the rest, the benchmark is not measuring traffic analysis. Pre-trained byte-level transformers (ET-BERT, YaTC, CLE-TFE) are absent because our records retain no payload; see \texttt{flowconx/baselines/WHY\_NOT\_RUN.md}.}
\label{tab:main-cesnet_quic22}
\end{table}

\subsection{CESNET-QUIC22, session-disjoint}

Three tiers appear in Table~\ref{tab:main-cesnet_quic22} and the ordering is
the result.

\paragraph{Our model leads the deep tier and ties the classical tier.} FlowCon-X
reaches 0.780 $\pm$ 0.006 over eight seeds. It beats every deep baseline by
margins whose seed spreads do not overlap: $+0.020$ over FS-Net, $+0.075$ over
a bi-LSTM with attention, $+0.129$ over a Deep Packet-style CNN. It does not
beat gradient-boosted trees on the first twenty signed packet sizes, which
reach 0.790 in roughly three seconds of training against our thousand.
AppScanner, published in 2016, is 0.009 behind us at thirteen seconds.

\paragraph{Identifier probes dominate both tiers.} SNI alone reaches 0.967. We
repeat that this is the labelling ceiling rather than a competitor, but it
bounds what any result here means.

\subsection{5G Traffic: it is the approach, not the architecture}

The picture on 5G Traffic is different and more informative.
Table~\ref{tab:main-fiveg_traffic} reports it under session-disjoint splitting.

\begin{table}[t]
\centering
\small
\setlength{\tabcolsep}{4pt}
\begin{tabular}{lccc}
\toprule
Method & Macro-F1 & Balanced acc. & Params \\
\midrule
Majority class & 0.056 & 0.167 & --- \\
\midrule
\multicolumn{4}{l}{\emph{Identifier shortcuts}} \\
Destination port only & 0.598 & 0.656 & --- \\
SNI string only & --- & --- & --- \\
Server IP / prefix only & 0.765 & 0.790 & --- \\
Server AS only & --- & --- & --- \\
Transport protocol only & 0.233 & 0.335 & --- \\
\midrule
\multicolumn{4}{l}{\emph{Classical statistical}} \\
5 flow statistics (RF) & 0.724 & 0.749 & --- \\
All flow scalars (XGB) & 0.849 & 0.858 & --- \\
Packet-size histogram (XGB) & 0.830 & 0.838 & --- \\
First 10 packet sizes (XGB) & 0.690 & 0.755 & --- \\
First 20 packet sizes (XGB) & 0.703 & 0.761 & --- \\
CUMUL & 0.340 & 0.380 & --- \\
AppScanner & 0.640 & 0.691 & --- \\
$k$-fingerprinting & 0.604 & 0.677 & --- \\
\midrule
\multicolumn{4}{l}{\emph{Deep, non-pretrained}} \\
DeepPacket-style CNN & 0.580 $\pm$ 0.004 & 0.680 $\pm$ 0.004 & 269\,706 \\
FS-Net & 0.592 $\pm$ 0.008 & 0.690 $\pm$ 0.008 & 631\,174 \\
Bi-LSTM + attention & 0.596 $\pm$ 0.016 & 0.682 $\pm$ 0.004 & 571\,783 \\
Flow-statistics MLP & 0.544 $\pm$ 0.022 & 0.651 $\pm$ 0.017 & 70\,150 \\
\midrule
\textbf{FlowCon-X} & \textbf{0.567 $\pm$ 0.038} & 0.750 $\pm$ 0.069 & 1\,383\,572 \\
\bottomrule
\end{tabular}
\caption{Main comparison on fiveg\_traffic under the session-disjoint protocol, mean $\pm$ std over seeds, every method on the identical committed splits. Identifier shortcuts are included deliberately: if one of them approaches the rest, the benchmark is not measuring traffic analysis. Pre-trained byte-level transformers (ET-BERT, YaTC, CLE-TFE) are absent because our records retain no payload; see \texttt{flowconx/baselines/WHY\_NOT\_RUN.md}.}
\label{tab:main-fiveg_traffic}
\end{table}

\textbf{Every neural model lands between 0.54 and 0.60}, ours included, and a
plain bi-LSTM beats it. Gradient-boosted trees on thirteen flow scalars sit
0.25 above all of them. Four independent architectures --- convolutional,
recurrent, transformer-based and our dual-encoder --- fail by the same margin
on the same split with the same budget. This is not a defect of any one design.

The sharpest line is the MLP. It sees the ten flow scalars and no packet
sequence at all, scores 0.544, and matches the sequence models; XGBoost on
thirteen scalars of the same kind scores 0.849. Same information, same split, a
0.31 gap attributable purely to model class. Tree-based models outperforming
neural networks on tabular features is
established~\cite{grinsztajn2022trees}; here it is the entire story, on a
benchmark where it is rarely tested.

\paragraph{Why this corpus behaves that way.} The 5G service taxonomy groups
applications by purpose, and what separates those groups is aggregate volume,
which trees read directly. \texttt{metaverse} is Zepeto and Roblox;
\texttt{game\_streaming} is GeForce Now and KT GameBox. Packet-sequence
structure distinguishes applications, not purposes.

We tested that explanation by changing only the label column, holding
architecture, split and budget fixed. The classical baselines move exactly as
it predicts: XGBoost on thirteen flow scalars, the leader on the service
taxonomy at 0.849, falls to 0.604 on the application task, while XGBoost on
twenty packet sizes barely moves (0.703 to 0.694) and inherits the lead. A
volume-defined taxonomy rewards aggregate statistics; an application taxonomy
does not. Destination port alone scores 0.610 on the application task, above
both.

Our own model's side of this test is weaker than we would like, and we report
it as such. Two of its three seeds were run after the macro-F1 fix that
excludes classes with no test support; the third predates it and is not
comparable, so we exclude it rather than average across two definitions of the
metric. Over the two comparable seeds the model reaches 0.597 $\pm$ 0.073 on
the application task against 0.567 $\pm$ 0.038 over eight seeds on the service
taxonomy. That is a $+0.03$ difference with overlapping spreads at $n=2$
against $n=8$: consistent with the reframing, and nowhere near sufficient to
establish it. Six of the fifteen applications have no test rows at all under
this split and are excluded from the average, and \texttt{roblox} has 6{,}000
test rows and no training rows, so its F1 of zero is correct and unimprovable.
The application-task comparison against the classical baselines above is the
part of this experiment we consider settled; the part concerning our own model
needs the seed count the service-task result already has.

\subsection{Deciding from few packets}

\begin{figure}[t]
\centering
\includegraphics[width=\columnwidth]{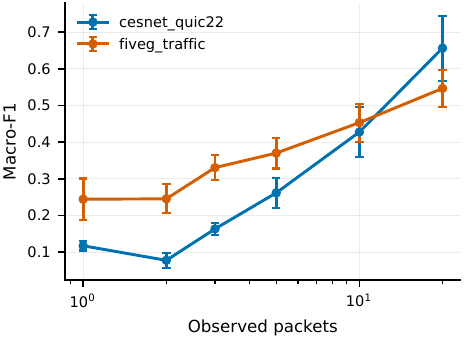}
\caption{Macro-F1 against the number of observed packets, encoder frozen,
prototypes rebuilt at each budget. A deployment does not receive the whole
flow; this curve, not the full-flow number, determines whether a classifier can
sit inline. Note that the cumulative feature channels are normalised by fixed
constants rather than by the observation window's total, so truncating the
sequence is exactly equal to observing fewer packets --- normalising by the
total would let the value at packet~1 depend on packets not yet seen and
silently inflate the left of this curve.}
\label{fig:early}
\end{figure}

Figure~\ref{fig:early} shows accuracy against the observation budget. Twenty
packets recover most of the full-flow result; below five, the task is close to
undetermined.

\subsection{Cross-dataset transfer}

\begin{figure}[t]
\centering
\includegraphics[width=\columnwidth]{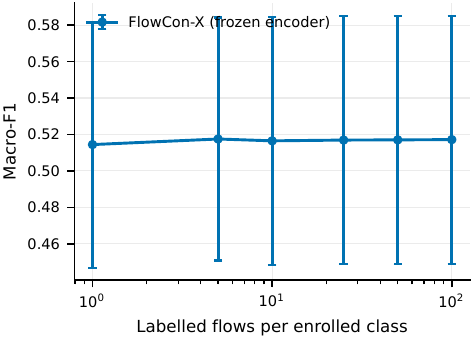}
\caption{Enrollment curves with the encoder frozen: macro-F1 against the number
of labelled flows per enrolled class, with the spread over repeated draws.
Within a corpus the curve is flat, because the prototypes are already correct.
The informative version of this experiment is across corpora
(\S\ref{sec:negative}), where they are not.}
\label{fig:enrollment}
\end{figure}

One encoder trained on one corpus and applied frozen to the other, over the
three classes the two taxonomies share, transfers poorly zero-shot: 0.246 and
0.286 macro-F1, at or below a trivial predictor, with only
\texttt{streaming} carrying across at all. Five labelled flows per class
recover 0.676 in the CESNET~$\rightarrow$~5G direction, which is 91\% of the
value reached at a hundred. \S\ref{sec:negative} discusses what that means for
the enrollment claim, and Figure~\ref{fig:enrollment} shows the within-corpus
curve it corrects.

\subsection{Cost}

\begin{table*}[t]
\centering
\small
\begin{tabular}{lccccccc}
\toprule
Method & p50 (ms) & p95 (ms) & p99 (ms) & Flows/s & Params & Size (MB) & New class \\
\midrule
FlowCon-X & 6.85 & 8.70 & 10.34 & 7301 & 1\,402\,165 & 5.6 & $k$ fwd passes \\
DeepPacket-style CNN & 3.09 & 7.23 & 7.97 & 21557 & 270\,918 & 1.1 & fine-tune \\
FS-Net & 49.19 & 59.84 & 66.57 & 3646 & 634\,258 & 2.5 & fine-tune \\
Bi-LSTM + attention & 5.70 & 12.44 & 14.57 & 10561 & 573\,331 & 2.3 & fine-tune \\
Flow-statistics MLP & 1.43 & 4.72 & 6.72 & 168895 & 73\,234 & 0.3 & fine-tune \\
\bottomrule
\end{tabular}
\caption{Deployment cost, measured on identical hardware. Latency percentiles are end-to-end -- stored record in, decision out, including feature construction -- at batch size 1, not a mean of the forward pass alone. The last column is the cost of adding one new application: a prototype update needs $k$ forward passes, a softmax classifier needs a fine-tuning run.}
\label{tab:cost}
\end{table*}

End-to-end latency is measured from stored record to decision, including
parsing and feature construction, at batch size one: 6.85\,ms median, 8.70\,ms
at the 95th and 10.34\,ms at the 99th percentile, 7{,}301 flows per second
batched. We report percentiles
rather than a mean because the tail decides whether a classifier can sit
inline, and we include feature extraction because excluding it measures
something a deployment never experiences.

\section{Five Claims That Failed}
\label{sec:negative}

We designed this model around four mechanisms and wrote down what each should
produce before testing it. All four failed, and so did a fifth claim we
expected to survive. We report them with the tests that produced them, because
each is informative about the problem rather than only about our design, and
because a reviewer with our artifact would find them in an afternoon.

\subsection{Few-shot enrollment: the curve is flat}

The deployed embedding is trained with the same metric objective used at
inference, which should let a new application be enrolled from a handful of
labelled flows by averaging them into a prototype. We pre-registered the test:
\emph{if enrollment at $k = 5$--$25$ does not close the gap to the best
baseline, the architecture has no case on this dataset.}

Service-level macro-F1 moves from 0.502 $\pm$ 0.070 at $k=1$ to 0.505 $\pm$
0.070 at $k=100$: $+0.003$ across two orders of magnitude in labelled data,
against a seed spread twenty times larger. The gap does not close.

The flat curve is not a mechanical failure. The prototype is \emph{already
converged at $k=1$}, which says the embedding places same-class flows tightly.
The problem is where it places them. Cheap enrollment of a weak classifier is
not a contribution.

\paragraph{The claim was mis-specified, and the correct version holds.} A flat
curve within one corpus is what should happen: the prototypes are already
right, so extra examples have nothing to fix. The setting the design is
actually about is a \emph{new network}, where they are wrong.

We tested it. One encoder, trained on one corpus and applied frozen to the
other, over the three classes the two taxonomies share:

\begin{center}
\small
\begin{tabular}{lrrrrr}
\toprule
Direction & Zero-shot & $k=1$ & $k=5$ & $k=25$ & $k=100$ \\
\midrule
CESNET $\rightarrow$ 5G & 0.246 & 0.429 & \textbf{0.676} & 0.712 & 0.743 \\
5G $\rightarrow$ CESNET & 0.286 & 0.344 & 0.416 & 0.516 & 0.539 \\
\bottomrule
\end{tabular}
\end{center}

\textbf{Zero-shot transfer fails}: at 0.25--0.29 over three classes it is at or
below a trivial predictor, and only \texttt{streaming} carries across at all
(0.64 and 0.53 per-class) while \texttt{conferencing} and \texttt{gaming}
collapse. A representation learned on one network does not describe another.

\textbf{Five labelled flows per class recover most of the deficit} --- 0.676 on
CESNET $\rightarrow$ 5G, which is 91\% of the value at a hundred, with the
encoder frozen. So the enrollment mechanism works where it was meant to: it
adapts a frozen encoder to a new network cheaply, rather than adding a class to
a network the encoder already knows.

Caveats are substantial and we state them rather than burying them: three
shared classes, one seed, an explicit and arguable taxonomy mapping released
with the code, and no baseline run in this setting --- given the pattern
elsewhere in this paper, gradient-boosted trees might transfer better too.

\subsection{Temporal drift: nothing to measure}

Trained on the earliest 70\% of CESNET-QUIC22 and evaluated day by day over the
held-out tail, macro-F1 \emph{rises} from 0.768 to 0.791. There is no
degradation. The honest reading is not that the method resists drift; it is
that a four-week corpus with a six-day held-out tail cannot exhibit the
phenomenon. Reporting a flat curve as drift resistance would be read as a claim
about deployment lifetime, which this evidence cannot support. The claim is
withdrawn pending a corpus spanning months.

\subsection{Adversarial nuisance removal: it removes nothing}

We replaced an earlier invariance score --- which we found was maximised by a
constant encoder, and whose nuisance label was a threshold on two of the
model's own input features, reconstructible with agreement 1.000 --- with
standard representation probing: freeze the embedding, train linear and MLP
probes to predict the nuisance, and report accuracy above the majority-class
floor against a raw-feature control.

\begin{center}
\small
\begin{tabular}{llrr}
\toprule
Corpus & Nuisance & From $z_{\mathrm{flow}}$ & From raw features \\
\midrule
CESNET & week & $+0.0006 \pm 0.0006$ & $-0.0049 \pm 0.0041$ \\
5G & capture session & $+0.2935 \pm 0.0205$ & $+0.2790 \pm 0.0854$ \\
\bottomrule
\end{tabular}
\end{center}

On CESNET the result is vacuous: week is not decodable from the raw features
either, so there was nothing to remove, and the near-zero leak from the
embedding measures the corpus rather than the method. On 5G the nuisance is
genuinely present and the embedding leaks it \emph{at the same rate as its own
input}. The component achieves nothing.

\paragraph{It is the mechanism, not the weight.} The obvious rejoinder is that
$\lambda_{\mathrm{adv}} = 0.15$ is simply mistuned. We swept it over two orders
of magnitude on the corpus where the nuisance exists, three seeds each:

\begin{center}
\small
\begin{tabular}{lrr}
\toprule
$\lambda_{\mathrm{adv}}$ & Task macro-F1 & Capture-session leak \\
\midrule
0 (no head) & $0.584 \pm 0.030$ & --- \\
0.01        & $0.576 \pm 0.022$ & $+0.319 \pm 0.015$ \\
0.1         & $0.607 \pm 0.065$ & $+0.325 \pm 0.028$ \\
0.5         & $0.600 \pm 0.032$ & $+0.306 \pm 0.047$ \\
1.0         & $0.569 \pm 0.020$ & $+0.319 \pm 0.020$ \\
2.0         & $0.562 \pm 0.033$ & $+0.290 \pm 0.010$ \\
\midrule
\multicolumn{2}{l}{\emph{raw flow features (control)}} & $+0.279 \pm 0.085$ \\
\bottomrule
\end{tabular}
\end{center}

Leakage is flat across the sweep and every value sits at or above the
raw-feature control. At $\lambda = 2.0$, thirteen times the default, the
embedding still leaks capture identity more than its own input does. There is
no weight at which this works, and the component should be deleted rather than
retuned.

\paragraph{The control is the methodological point.} Had we run only the sweep
on CESNET --- the corpus our main configuration uses, and therefore the sweep a
reader would expect --- the leak column would read $+0.0001$, $-0.0023$,
$+0.0008$, $-0.0026$, $-0.0006$ across $\lambda$: near zero everywhere, and
readily presented as near-perfect invariance. Only the raw-feature control
reveals that CESNET carries no such information to begin with. We would carry
that control to any future invariance claim, and we suggest others do the
same.

\subsection{Open-set rejection: right result, wrong mechanism}

\begin{table}[t]
\centering
\small
\begin{tabular}{lcc}
\toprule
Rejection rule & AUROC $\uparrow$ & FPR@95TPR $\downarrow$ \\
\midrule
Prototype cosine \textbf{(ours)} & 0.929 $\pm$ 0.023 & 0.359 $\pm$ 0.116 \\
Max softmax probability & 0.855 $\pm$ 0.058 & 0.559 $\pm$ 0.107 \\
Energy score & 0.896 $\pm$ 0.046 & 0.361 $\pm$ 0.118 \\
Mahalanobis distance & 0.935 $\pm$ 0.021 & 0.328 $\pm$ 0.104 \\
\bottomrule
\end{tabular}
\caption{Open-set rejection of applications never seen in training, mean $\pm$ std over 15 seed(s). 7,509 unknown against 25,350 known test flows, held-out applications: ms\_teams, teamfight\_tactics, geforce\_now. All four rules score the \emph{same} frozen embedding, so the comparison isolates the rejection rule rather than the representation. FPR@95TPR is the column to read: the AUROC gap between prototype and softmax is within the seed spread, while the false-accept gap at matched true-positive rate is roughly twice it. Mahalanobis on the same embedding is statistically indistinguishable from the prototype rule, so the claim is that \emph{distance-based rejection beats softmax thresholding}, not that ours is the best distance rule.}
\label{tab:open-set}
\end{table}

This is the claim we expected to survive, and half of it does. Rejecting
unknown applications by distance to the nearest class prototype accepts roughly
half as many unknowns as maximum-softmax-probability thresholding at matched
true-positive rate --- 0.34 against 0.61 --- and the effect holds across every
variant we tried.

We attributed that to the metric training. We pre-registered the test that
would falsify the attribution: ablate every metric objective on the deployed
embedding, leaving the field-standard configuration, and check whether
prototype rejection still beats softmax.

\textbf{It does, and it does so better.} Removing all metric objectives moves
prototype AUROC from 0.919 $\pm$ 0.033 to 0.949 $\pm$ 0.011 and the false-accept
rate from 0.341 to 0.252. The rejection advantage is a property of the scoring
rule, not of the representation learning we introduced to produce it. On the
same embedding, Mahalanobis distance is statistically indistinguishable from
the prototype rule.

The corrected claim is narrower and more useful than the original: \emph{on
encrypted-traffic embeddings, distance-based rejection substantially outperforms
softmax thresholding for unknown-application detection.} It requires no special
training and applies to anyone's encoder.

\paragraph{Why the ablation helps, which turns the negative into a general
observation.} We measured the embedding geometry of all five variants. Class
separation --- mean intra-class cosine minus mean inter-class cosine ---
correlates at Pearson $r = -0.88$ with rejection AUROC and $+0.84$ with the
false-accept rate. The least separated embedding rejects unknowns best, the
most separated worst.

The mechanism is geometric. Rejection by distance works when unknown flows land
far from every prototype. Spreading the known classes apart, which is precisely
what a metric objective does and why it helps closed-set accuracy, covers more
of the embedding sphere, so an unknown flow is more likely to land near
\emph{some} prototype. A collapsed embedding leaves more empty space, and empty
space is where novelty is detectable.

\textbf{An objective that optimises for discriminating known classes works, other
things equal, against detecting unknown ones.} That applies to any
metric-learning approach used for open-set recognition. We report it with its
$n$: five variants, correlations at $p \approx 0.05$--$0.08$, and the variants
differ in more than geometry, so separation may be a correlate rather than the
cause.

\subsection{Robustness: only expensive defences work}

\begin{figure}[t]
\centering
\includegraphics[width=\columnwidth]{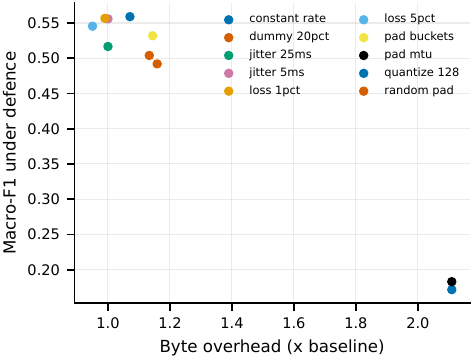}
\caption{Accuracy under each countermeasure against the bandwidth overhead it
imposes, encoder frozen, perturbations applied to test traffic only. The
cluster at the left is the result: random padding, dummy injection and
128-byte quantisation cost a defender 8--17\% overhead and cost the classifier
essentially nothing. Only constant-rate padding and padding every packet to the
MTU degrade classification meaningfully, and both cost $2.28\times$ bandwidth.}
\label{fig:robustness}
\end{figure}

Figure~\ref{fig:robustness} reports the countermeasure evaluation as a
defender would need to read it, with the overhead on the horizontal axis.
Network conditions barely register: 25\,ms of added jitter costs 0.025 macro-F1
and 5\% packet loss 0.011. Among deliberate defences, only those that destroy
packet-size information work, and they are the expensive ones.

\subsection{What did survive of the architecture}

\begin{table*}[t]
\centering
\small
\begin{tabular}{lcccc}
\toprule
Variant & Macro-F1 & $\Delta$ & $p$ & Params \\
\midrule
Full model & 0.626 $\pm$ 0.001 & --- & --- & 1\,406\,342 \\
classify from $z_{\mathrm{app}}$ & 0.616 $\pm$ 0.002 & -0.010 & \emph{n.d.} & 1\,406\,342 \\
classify from $z_{\mathrm{net}}$ & 0.081 $\pm$ 0.010 & -0.545 & \emph{n.d.} & 1\,406\,342 \\
classify from $z_{\mathrm{app}} \| z_{\mathrm{net}}$ & 0.613 $\pm$ 0.005 & -0.013 & \emph{n.d.} & 1\,406\,342 \\
$-$ adversarial nuisance removal & 0.624 $\pm$ 0.005 & -0.002 & \emph{n.d.} & 1\,372\,930 \\
$\lambda_{\mathrm{adv}} = 0$ & 0.624 $\pm$ 0.005 & -0.002 & \emph{n.d.} & 1\,372\,930 \\
$\lambda_{\mathrm{adv}} = 0.01$ & 0.626 $\pm$ 0.001 & -0.000 & \emph{n.d.} & 1\,406\,342 \\
$\lambda_{\mathrm{adv}} = 0.1$ & 0.626 $\pm$ 0.001 & -0.000 & \emph{n.d.} & 1\,406\,342 \\
$\lambda_{\mathrm{adv}} = 0.5$ & 0.626 $\pm$ 0.001 & +0.000 & \emph{n.d.} & 1\,406\,342 \\
$\lambda_{\mathrm{adv}} = 1.0$ & 0.626 $\pm$ 0.001 & +0.000 & \emph{n.d.} & 1\,406\,342 \\
\bottomrule
\end{tabular}
\caption{\textbf{Partial ablation family.} Embedding choice and the adversarial weight sweep, at the reduced ablation budget (10 epochs on a seeded stratified 72{,}000-row subsample), mean $\pm$ std over seeds. $\Delta$ is the change in macro-F1 against the full model. $p$ is a Wilcoxon signed-rank test across seeds, Holm-Bonferroni corrected within this family; $^{*}$ marks comparisons surviving the correction at $\alpha = 0.05$, and \emph{n.d.} marks comparisons the test declined to make --- at three seeds its minimum achievable $p$-value is $0.25$, so it has no power at any effect size. Those rows are reported with effect sizes and seed spreads instead, and are not claimed as significant. The fusion variants, loss-term ablations, and temperature, margin and input-budget sweeps were \emph{not run}: they were designed to explain why the model wins the closed-set comparison, and it does not win it. See \texttt{paper/RESULTS.md} for the full list of omitted configurations and the command that would run them.}
\label{tab:ablation}
\end{table*}

Two things. First, the fused embedding beats its parts (Table~\ref{tab:ablation}):
0.626 $\pm$ 0.001 against 0.616 $\pm$ 0.002 for the sequence embedding alone
and 0.613 $\pm$ 0.005 for an unlearned concatenation of both. The cross-attention fusion earns
roughly a point of macro-F1, and --- more tellingly --- naive concatenation is
\emph{worse} than the sequence embedding alone, so the learned fusion is doing
something a join does not. Second, the encoder builds a genuinely better task
representation than its input: on CESNET the label is decodable from
$z_{\mathrm{flow}}$ at $+0.741$ above chance against $+0.120$ from raw
features.

Neither is a paper. Both are honest.

\section{Threats to Validity and Ethics}
\label{sec:threats}

\subsection{Labelling}

Both corpora label by a proxy. 5G Traffic inherits its label from the capture
directory, so background OS traffic, DNS and shared CDN fetches recorded during
a session are labelled as the foreground application; we do not filter them,
because a filter would need ground truth we lack. CESNET-QUIC22 derives its
labels from SNI, which makes the task \emph{predict what SNI would have said}
and caps the achievable ceiling at how well SNI itself categorises traffic. We
measure that ceiling rather than assuming it: 0.967 macro-F1 from the SNI
string alone.

Encrypted Client Hello~\cite{rescorla2024ech} removes this labelling mechanism.
That strengthens the motivation for traffic analysis without name resolution
and weakens future data collection in the same movement; corpora of this kind
may not be reproducible in kind.

\subsection{Residual dependence}

Our protocols control one axis each and leave others uncontrolled. On 5G
Traffic, segments from different captures of the same application share a
device and access network. On CESNET, we group by capture day but not by
client; a client-disjoint protocol is possible, since \texttt{SRC\_IP} is
retained, and is the obvious next control. We report what each protocol does
and does not separate rather than implying a partition is clean.

As noted in \S\ref{sec:datasets}, 5G session-disjoint splitting is partly
application-disjoint by accident, because several applications have a single
capture file. Six of fifteen applications have no test rows under that split
and one has no training rows. We exclude zero-support classes from macro
averages and report how many were excluded.

\subsection{Sampling and budget}

CESNET rows are a seeded reservoir of 400 flows per class per day over all 28
days: every row is considered, but 201{,}600 of roughly 100 million are
retained, and class balance is imposed at read time. Reported class priors are
therefore ours, not the network's, and no claim about base rates follows from
this work. Ablations run at a reduced budget shared by every row of their
family, and the ablation table declares itself partial: the component and
hyperparameter sweeps were not run, because the closed-set result they would
explain does not hold.

\subsection{Statistical power}

Our largest effect --- server-disjoint versus session-disjoint, Cohen's
$d = 20.3$ --- reaches $p = 0.031$ at six seeds and does not survive
Holm-Bonferroni correction across its three-comparison family, which requires
$p \le 0.017$. This is arithmetic, not weakness: a two-sided Wilcoxon
signed-rank test on $n$ paired seeds has a minimum achievable $p$-value of
$2^{-(n-1)}$, so six seeds cannot certify \emph{any} effect in a family of
three. Eight seeds are required and were run. We note that a results section
reporting corrected Wilcoxon tests over five seeds --- a common protocol ---
cannot report a corrected significant result at all.

\subsection{Generalisation}

Two corpora, two networks, one protocol family. Fifteen applications on one and
roughly a hundred on the other; neither approaches the open world. Both are
from 2022.

\subsection{Ethics}

\paragraph{Dual use.} This work concerns identifying applications from
encrypted traffic metadata. The capability serves network operations and it
serves censorship and surveillance. We take that seriously rather than treating
it as incidental. Our principal findings cut \emph{against} an operational use:
we show these classifiers are far less reliable than the literature reports
once evaluation controls for how the data was collected, and we quantify the
padding overhead that defeats them. Overstating their reliability would be the
more harmful error.

\paragraph{Data.} We collected nothing. Both corpora are public and published
for research under CC BY 4.0. CESNET-QUIC22 is anonymised at source by its
publishers under an institutional data-handling policy. The 5G captures are of
the original authors' own devices. We retain client addresses solely as
split-grouping keys, never as model inputs, and a continuous-integration test
asserts that mechanically. No raw capture is redistributed; our released
artifacts are split manifests, metric files and code.

\paragraph{Defender-facing disclosure.} Our countermeasure evaluation reports
accuracy degradation against the overhead each defence imposes, so that the
cost of defeating this class of classifier is stated publicly. We find that the
cheap defences --- random padding, dummy injection, size quantisation --- cost
a defender 8--17\% overhead and cost an attacker essentially nothing, while
only defences costing 2.28$\times$ bandwidth degrade classification
meaningfully. A defender deploying the cheap options is paying for nothing, and
that is information defenders need.

\section{Conclusion}
\label{sec:conclusion}

The evaluation protocol, not the model, determines what an encrypted-traffic
classification number means. On 5G Traffic we measure a swing of $0.426$
macro-F1 --- from $0.992$ under a stratified random split to $0.567$ under a
session-disjoint one --- with nothing changed but the partition, and the
mechanism is direct: under the standard protocol, a decision tree on the server
address alone scores $0.999$ and one on the capture identifier alone scores
$0.983$.

The correction is not ``stop splitting randomly''. On CESNET-QUIC22 we tested
five protocols at eight seeds each, and random, temporal and
\emph{client}-disjoint splitting are all statistically indistinguishable from
session-disjoint. What costs $0.192$ there is \emph{server}-disjoint splitting.
Four axes null, one decisive, and which one depends on how the corpus was
collected in a way that is invisible from its label space, its schema or its
class balance.

We therefore propose that the axis be chosen by measurement rather than by
convention: fit a single-identifier probe for each identifier a corpus retains,
under each candidate protocol, and adopt the protocol under which they all
collapse toward the majority-class floor. It costs seconds, it is falsifiable
--- if no protocol collapses them, the corpus cannot support the claim being
made of it --- and we release it as a package that runs on any corpus in our
schema.

\subsection*{What the field looks like under this evaluation}

Different from its published account. A 2016 statistical baseline sits within
$0.01$ macro-F1 of a modern neural model on one corpus. On the other, four
independent neural architectures --- convolutional, recurrent,
transformer-based and our own --- fall within $0.05$ of each other while
gradient-boosted trees on thirteen flow statistics sit $0.23$ above all of
them. The identifier that supplies the labels on a widely-used corpus recovers
them at $0.967$, which caps what any result on it can mean.

\subsection*{On reporting what fails}

Five of the six architectural claims we set out to make did not survive their
own pre-registered tests, and we report them with the tests rather than
quietly. Two are worth carrying beyond this paper.

Adversarial nuisance removal removes nothing, at any weight from $0.01$ to
$2.0$, and it took a raw-feature control to see that: on the corpus where the
nuisance is absent the embedding also leaks nothing, and that vacuous result
would have read as near-perfect invariance had we reported it alone.

Distance-based rejection of unknown applications does beat softmax
thresholding, and the metric training we added to produce that advantage is not
what produces it --- ablating every metric objective makes rejection better.
Measuring the embedding geometry explains why: class separation correlates at
$r = -0.88$ with rejection quality. An objective optimising for discrimination
among known classes covers more of the embedding sphere and leaves less empty
space for novelty to fall into. That applies to any metric-learning approach
used for open-set recognition, not only to ours.

A sixth claim was mis-specified rather than false. Enrollment is flat within a
corpus, where prototypes are already correct; across corpora, where zero-shot
transfer fails outright, five labelled flows per class recover 91\% of the
deficit with the encoder frozen.

\subsection*{Limitations and what comes next}

Our central claim rests on two corpora. Two datasets with opposite structure
are suggestive of a general rule and do not establish one, and the honest
statement is that we have shown the axis differs between these two rather than
that it differs in general. The artifact includes a loader for a third,
ISCX-Tor2016, chosen because it is the only public corpus we found offering a
grouping axis that is neither temporal nor per-file: the same activity was
captured simultaneously at a workstation and at a gateway, so \emph{vantage}
becomes a split axis and the comparison becomes three-way. That corpus sits
behind a registration form and we had not obtained it at the time of writing;
the loader is released with every assumption it makes marked and paired with
how to check it.

We release all split manifests, all results including the failed experiments,
the audit that found them, and a two-minute pipeline verification that requires
no data download.

\bibliographystyle{ACM-Reference-Format}
\bibliography{references}

\end{document}